\documentclass{article}

    \usepackage[preprint]{tackling_climate_workshop_style}

\usepackage[utf8]{inputenc} 
\usepackage[T1]{fontenc}    
\usepackage{hyperref}       
\usepackage{url}            
\usepackage{booktabs}       
\usepackage{amsfonts}       
\usepackage{nicefrac}       
\usepackage{microtype}      
\usepackage{natbib}
\usepackage[final]{graphicx}
\usepackage{subcaption}

\title{\textit{Continual VQA for Disaster Response Systems}}
  
\newcommand{\repeatthanks}{\textsuperscript{\thefootnote}}

\author{%
  Aditya Kane\thanks{Indicates equal contribution} \\
  Pune Institute of Computer Technology, Pune\\
  \texttt{adityakane1@gmail.com} \\
   \And
   V Manushree\repeatthanks \\
   Manipal Institute of Technology, Manipal \\
   \texttt{manushree635@gmail.com} \\
   \And
   Sahil Khose\repeatthanks\\
   Manipal Institute of Technology, Manipal \\
   \texttt{sahilkhose18@gmail.com} \\
}
    
\begin{document}

\maketitle

\begin{abstract}

    Visual Question Answering (VQA) is a multi-modal task that involves answering questions from an input image, semantically understanding the contents of the image and answering it in natural language. Using VQA for disaster management is an important line of research due to the scope of problems that are answered by the VQA system. However, the main challenge is the delay caused by the generation of labels in the assessment of the affected areas. To tackle this, we deployed pre-trained CLIP model, which is trained on visual-image pairs. however, we empirically see that the model has poor zero-shot performance. Thus, we instead use pre-trained embeddings of text and image from this model for our supervised training and surpass previous state-of-the-art results on the FloodNet dataset. We expand this to a continual setting, which is a more real-life scenario. We tackle the problem of catastrophic forgetting using various experience replay methods. \footnote{Our training runs are available at: \url{https://wandb.ai/compyle/continual_vqa_final}}

\end{abstract}




\section{Introduction}
Climate change is one of the most pressing issues of our times, with increased greenhouse gases emission and urbanization. One of the most destructive effects of climate change is flooding, disrupting people's lives and causing damage to properties. During floods, most disaster response systems rely on aerial data to assess the affected areas and provide immediate help. 

In applications, like a deployed disaster response system, where labels and data are going to be generated in real time and the model is deployed for predictions, the most important thing is to save time and deploy rescue teams as quick as possible based on the model predictions. There are going to be two types of major delays for such a system: labelling the entire data at once and training the model on the labeled data. The latter problem specific to FloodNet VQA was addressed by \cite{kane2022efficient}, which reduced the training time by 12x and inference time by 4x on the entire data. However, the former is yet not addressed, as labelling takes a lot of time. 

To address the delay caused by label generation, we investigate the direct evaluation of large pre-trained multimodal models on the downstream VQA dataset, studying the effects of zero-shot transfer, proposed by \citet{clipmodels2022songetal}. We analyze CLIP's zero-shot ability on the FloodNet VQA dataset. These models do not perform well on the downstream aerial VQA task without finetuning the entire dataset. We, therefore, use pre-trained CLIP embeddings of the images and the questions from these models for our supervised training. Previous works like \citet{hurmicvqa} solve this problem using VGG-19 and LSTM features, but their results are inferior to modern architectures like ResNet\cite{resnet} and BERT-based \cite{bert} models.

Our main contribution in this paper is to tackle this real-time setting delay with the help of continual learning for visual question answering by training our models on a continuous stream of data using experience replay. The models are trained on different tasks in a sequential fashion without substantial forgetting the previous tasks. This way, the need for complete data labelling is eliminated, reducing training time and compute requirements. We provide a thorough analysis of the task-order permutations and compare the same with supervised benchmarks.
\section{Methodology}

\label{sec:method}

We use the FloodNet dataset \cite{floodnet} for our experiments. It is a VQA dataset based on aerial images of flooded areas. We have elaborated the dataset in Appendix \ref{app:dataset}. In the ideal case, pretrained multimodal models like CLIP \cite{clip} should correctly answer questions related to image due to the shared latent space between image and text features. Bearing that in mind, we experiment with an out-of-the-box CLIP predictor to answer questions. Concretely, we evaluate the performance of CLIP on the FloodNet VQA task without finetuning. We convert the question text to prompts containing the probable labels for our text input. We then pass these inputs to the CLIP model, which returns the posterior probability of every possible label word. 

However, this ideal scenario is unlikely, more so in the case of specialized aerial images. The following best approach is task-wise lifelong learning. Here, the data is available through a stream of tasks, meaning a task once trained on cannot be trained on again. This more closely depicts the real-world scenario where the labels are not available all at once. Thus, we divide the available dataset into three tasks: Yes/No, Road condition recognition and Image condition recognition. We use experience replay \cite{chaudhry2019tiny} to further enhance the continual learning abilities of the model. Given that image and text features from models like CLIP are more coherent due to the shared latent space, we use these features for continual training. We use the training strategy proposed by \citet{kane2022efficient}. We also provide results on this model trained on the complete dataset and task-wise training results as the upper bound for the continual training setup.  

\section{Experiments and Results}

\begin{table*}[b]
\centering

\begin{tabular}{|c|cccc|}
\hline
\textbf{Method}            & \multicolumn{4}{c|}{\textbf{Taskwise Accuracy}}                                                                             \\ \hline
                  & \multicolumn{1}{c|}{\textbf{Overall}} & \multicolumn{1}{c|}{\textbf{Yes/No}} & \multicolumn{1}{c|}{\textbf{Image Condition}} & \textbf{Road Condition} \\ \hline
CNX-mul           & \multicolumn{1}{c|}{98.03}   & \multicolumn{1}{c|}{98.31}  & \multicolumn{1}{c|}{\textbf{98.62}}           & 97.18          \\ \hline \hline
CLIP-ZS           & \multicolumn{1}{c|}{35.56}   & \multicolumn{1}{c|}{15.12}  & \multicolumn{1}{c|}{41.72}           & 83.14          \\ \hline\hline
CLIP-add          & \multicolumn{1}{c|}{93.99}   & \multicolumn{1}{c|}{88.37}  & \multicolumn{1}{c|}{95.17}           & 97.67          \\ \hline
CLIP-cat          & \multicolumn{1}{c|}{92.97}   & \multicolumn{1}{c|}{81.97}  & \multicolumn{1}{c|}{95.86}           & 97.06          \\ \hline
CLIP-mul          & \multicolumn{1}{c|}{96.4}    & \multicolumn{1}{c|}{97.71}  & \multicolumn{1}{c|}{95.17}           & 97.14          \\ \hline
CLIP-mul-taskwise & \multicolumn{1}{c|}{\textbf{98.33}}   & \multicolumn{1}{c|}{\textbf{98.85}}  & \multicolumn{1}{c|}{98.43}           & \textbf{97.71}          \\ \hline

\end{tabular}

\caption{Supervised, Taskwise supervised and Zero shot results on CLIP features. We use the best result from \citet{kane2022efficient} as our baseline.}
\label{tab:final_results}
\end{table*}

\begin{figure}
\begin{subfigure}{.5\textwidth}
  \centering
  \includegraphics[width=.8\linewidth]{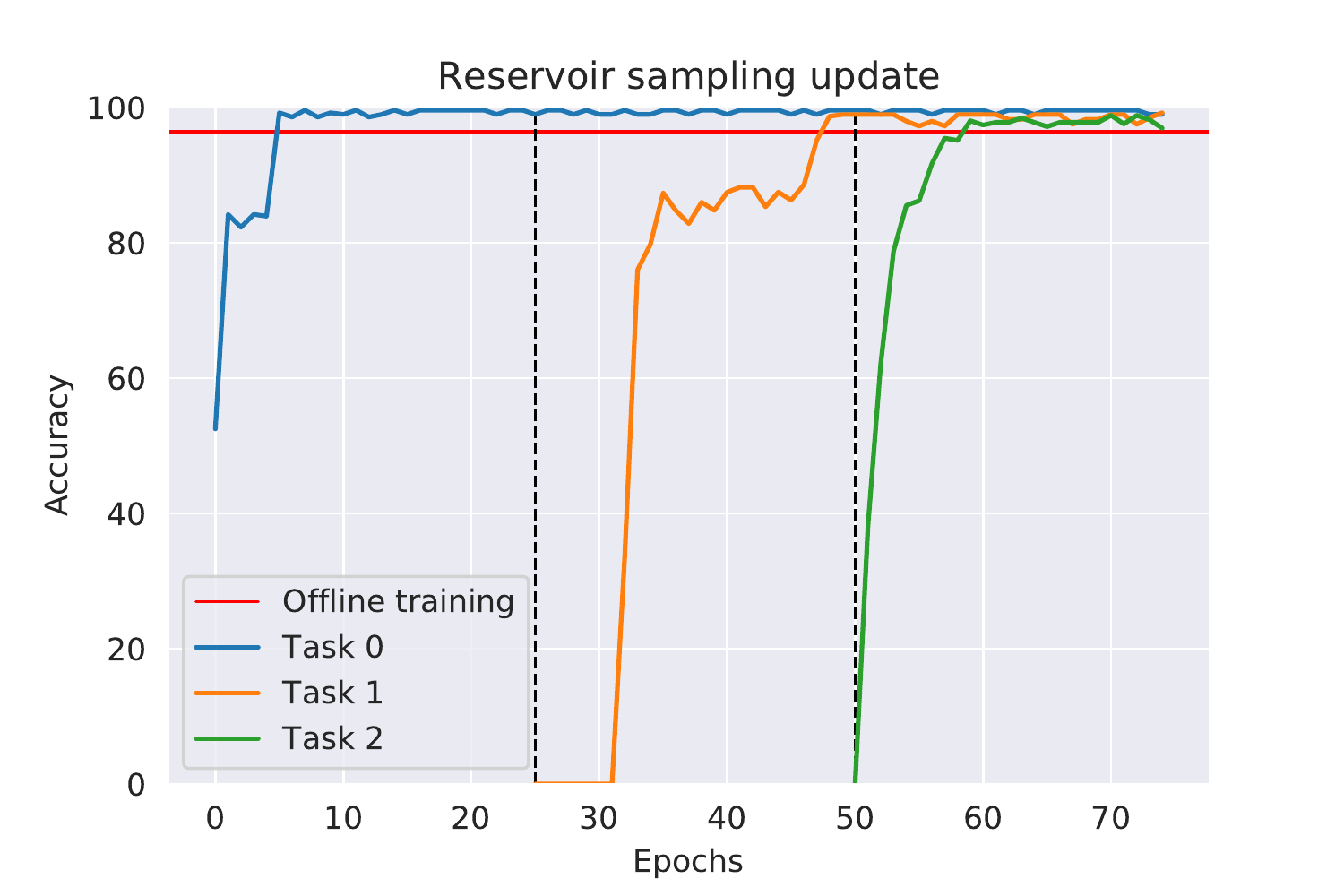}
  \caption{Reservoir Sampling Update}
  \label{fig:Reservoir Sampling Update}
\end{subfigure}%
\begin{subfigure}{.5\textwidth}
  \centering
  \includegraphics[width=.8\linewidth]{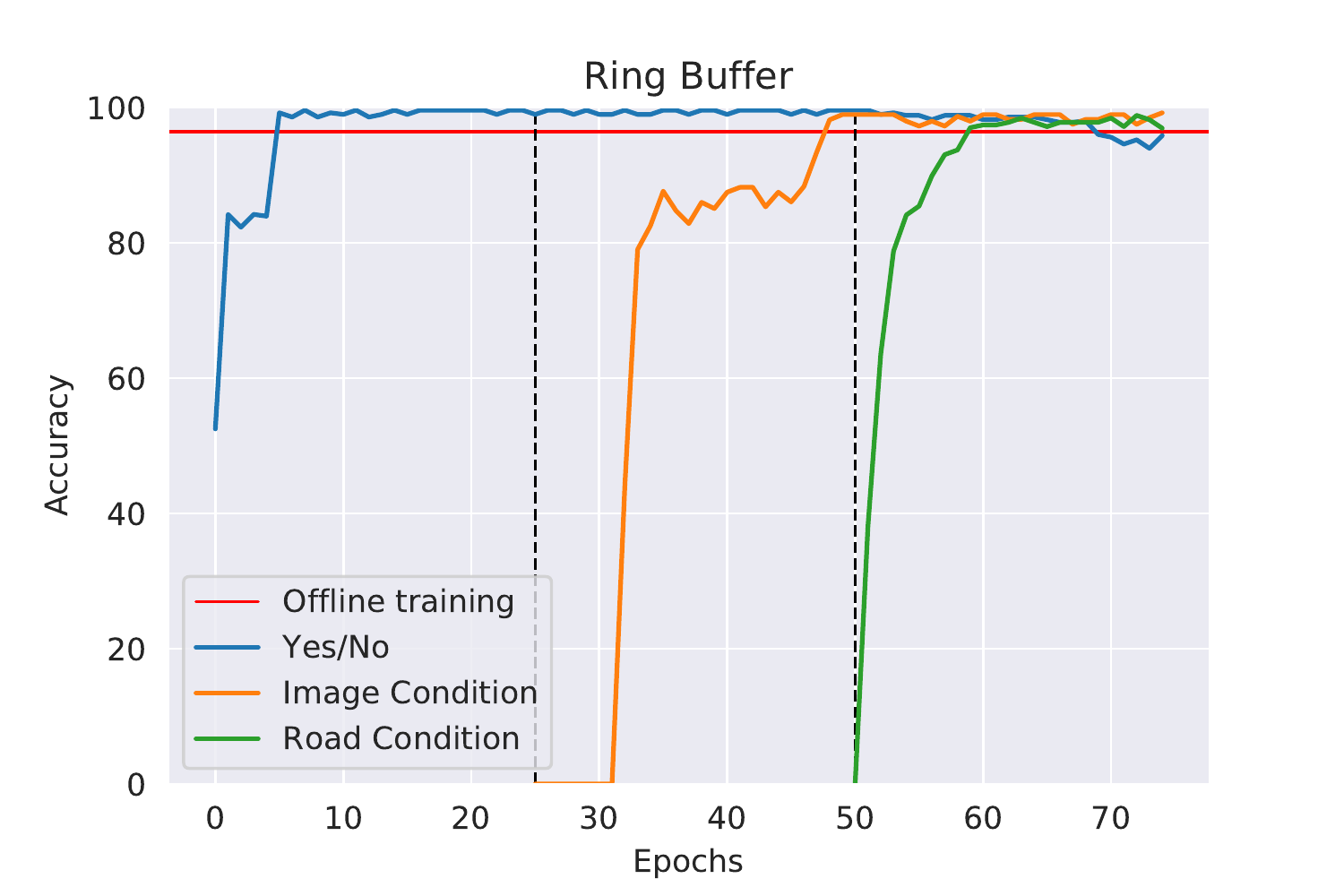}
  \caption{Ring Buffer}
  \label{fig:Ring Buffer}
\end{subfigure}
\centering
\begin{subfigure}[h]{.5\textwidth}
  \centering
   \includegraphics[width=.8\linewidth]{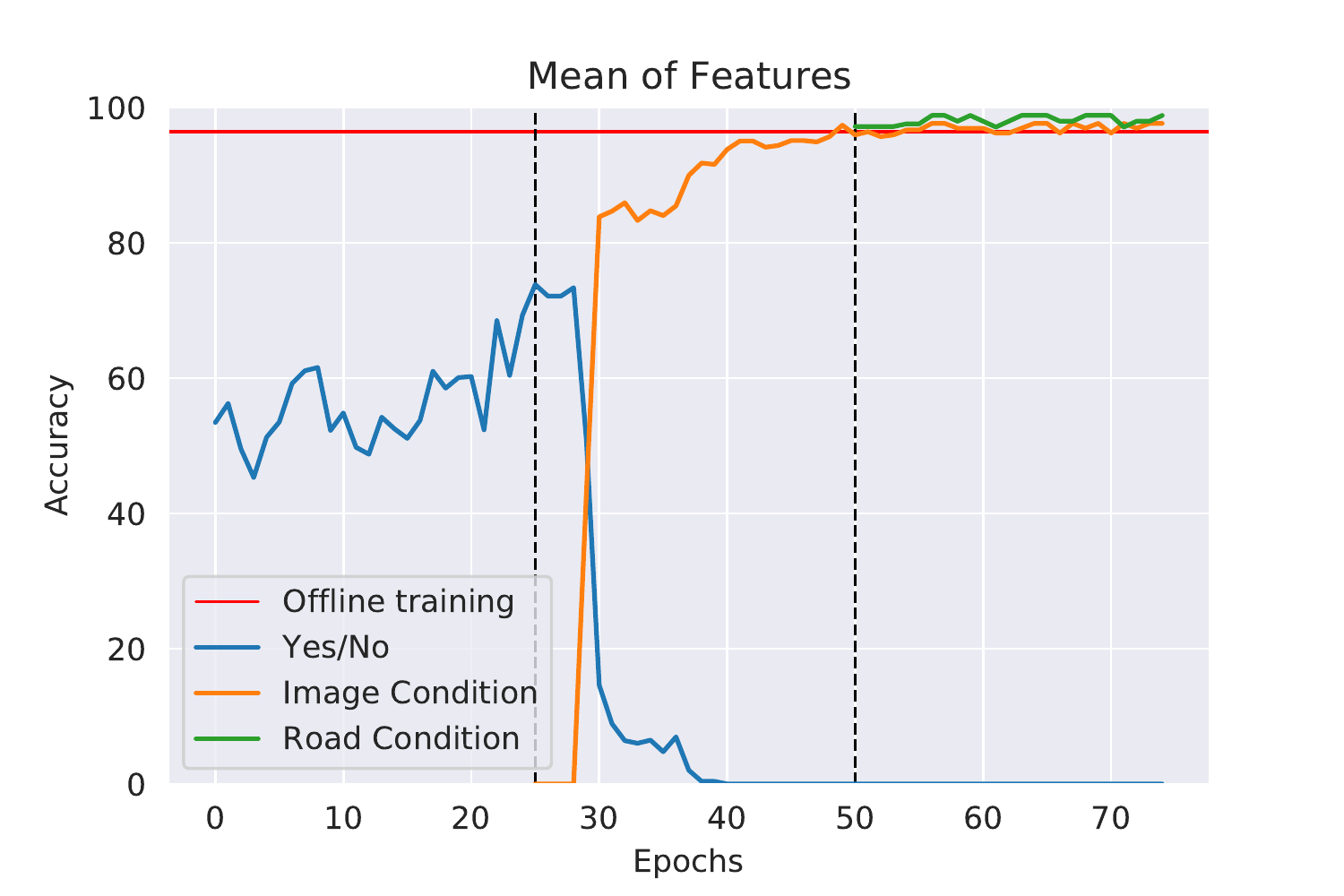}
    \caption{Mean of Features}
  \label{fig:Mean of Features}
\end{subfigure}%

\caption{Experience replay methods used in our continual setup for VQA on the FloodNet dataset}
\label{fig:main_results}
\end{figure}

\subsection{Zero shot setup}
 
We evaluate CLIP's zero shot ability on FloodNet VQA, and as suggested by \citet{clipmodels2022songetal} as a "first-attempt". The intuition behind this is to empirically establish if zero-shot methods are sufficient for VQA tasks like FloodNet. We use OpenAI's open source implementation of CLIP and use ViT L-14 \cite{vit} image encoder for evaluation. 
Since CLIP is a model that has only two input modalities, we cannot directly use both question and answers. Thus, we create prompts for all probable answers and then filter the answers based on the question. The model then predicts the most probable answer. The results of this experiment are mentioned in Table \ref{tab:final_results}. We see that this method performs poorly. 

\subsection{Continual setup and experiments}

As explained in Section \ref{sec:method}, the logical alternative for the zero shot experiments is the continual setup. We use the Experience Replay (ER) \cite{chaudhry2019tiny} based algorithm for training our model in continual fashion and compare it against disjoint model training for each of the individual tasks. This has shown to greatly help in reducing catastrophic forgetting. We observe the same in our experiments as well. In Figure \ref{fig:main_results}, we show various experience replay methods. We observe that they surpass full-dataset supervised benchmarks. We speculate that this is because of transfer learning between consecutive tasks. All possible permutations for various tasks are explored for the CL setup and the optimal order of tasks is proposed for optimal results. In a real-world scenario, the optimum task permutation might not be possible. We  provide the results of these permutations in Appendix \ref{app:permutations}. 

We observe that using CLIP features greatly improves task-wise training scores. In Table \ref{tab:final_results}, "CNX-mul" stands for ConvNeXt + RoBERTa features with multiplication combination, "CLIP-ZS" stands for zero-shot CLIP, "CLIP-\{add, mul, cat\}" stand for addition, multiplication or concatenation of CLIP-based features. Lastly, "CLIP-mul-taskwise" is same as "CLIP-mul" but each task is trained and evaluated separately.

We also observe that our system is robust to catastrophic forgetting. Our results are shown in Figure \ref{fig:main_results} denote the previous and current task accuracies while the model is trained on the current task. We observe that reservior sampling update performs the best amongst the three methods. Lastly, we use a four-stage, three-tier MLP architecture to train our system. We use this architecture to ensure that the model does not underfit, given all tasks.

\subsection{Results and observations}

As explained above, we perform extensive studies of zero shot, continual and supervised capabilities of our models. Our implementation details are available in Appendix \ref{app:impl}. We present several interesting observations for our experiments:

\begin{enumerate}
    \item \textbf{Out-of-the-box Zero-shot evaluation on CLIP performs poorly}: CLIP \cite{clip} performs poorly when evaluated out-of-the-box on our testing dataset. We speculate that this is because CLIP is not exposed to a substantial amount of aerial images. Fine-tuning CLIP on aerial images may lead to improved performance and can be viewed as potential future work.
    \item \textbf{Supervised training using CLIP features outperforms state-of-the-art on FloodNet VQA}: We find that using CLIP features instead of ConvNeXt + RoBERTa features improves overall performance of the model. We hypothesize that this is because CLIP features are in the same latent space as opposed to ConvNeXt and RoBERTa's features.
    \item \textbf{Reservoir sampling is the best memory update method}: We observe that memory update in experience replay outperforms other memory update method by a considerable margin. We attribute this to the presence of original samples in the buffer from all tasks as opposed to ring or mean-of-features method.
    \item \textbf{"Image Condition" and "Road Condition" tasks yield consistent results}: We see that "Image Condition" and "Road Condition" tasks often "complement" each other and their performance is almost equivalent across different task permutations. It can be seen that this is because both of these tasks use the same labels. Moreover, "Road Condition" can be said to be a subset of "Image Condition" task since the condition of the central road in an image is often the condition of an entire image. Thus, it can be hypothesized that these two tasks exhibit efficient transfer learning.
    \item \textbf{"Yes/No" task hampers performance}: The "Yes/No" task is seen to degrade performance, when trained before, after or between the condition recognition tasks. However, as mentioned in earlier point, some extent of transfer learning can be seen across this task as well, since most questions in this category pertain to the condition of the image or road. Still, we speculate that the performance degradation occurs due to inactivity of label neurons in the network other than those used for this task.
\end{enumerate}

\section{Addressing Impact}
In this paper, we try to build efficient systems to mitigate delay in assessing flood affected areas by the disaster response teams.  We develop a visual question answering system more suited for real-world settings, where label generation is a very time-consuming task. We  build a continual based system for our tasks, where our model learns on-the-fly on new tasks without re-training on the old tasks again. This way we tackle the delay in label generation, as the deployed models can learn continuously as the labels for new tasks are being generated. This way the flood disaster management and response system can help those in distress as soon as possible. The tasks order can be tailored, so the tasks which are more important for assessment and needs immediate attention can be trained before, instead of wasting precious time in these disaster situations on generating labels for all the tasks.

\section{Conclusion}
We explore the Continual VQA setup, which has been unexplored before, especially in disaster-response tasks like FloodNet. This training setup is meant to solve two problems in disaster response systems -- 1. Modern-VQA architectures are computationally expensive. We use the light-weight proposed model by \citet{kane2022efficient}, which mitigates the problem of model deployability, 2. It is not realistic to wait for the generation of all the labels on the entire dataset in disaster-response systems. We address this by equipping our light-weight model with online training capability and also suggest the order of task labels to be provided for optimal performance of the deployed model. This facilitates quicker deployment of the model for real-time response action and the capability of the model to learn in an online fashion after the availability of newer labels. 

\section*{Acknowledgement}

We greatly thank Neeraja Kirtane for her help with drafting and reviewing of this paper, without which this paper would not have been possible.

\bibliography{custom}
\bibliographystyle{acl_natbib.bst}


\appendix

\section{Dataset}
\label{app:dataset}
FloodNet introduces the VQA task for UAV-based disaster response system. The collected data from disaster-response systems are used to solve various problems.
\cite{khose2021semi} has proposed a semi-supervised solution for the classification and multi-class semantic segmentation problem by utilizing pseudo-label training on the massive amount of unlabeled samples in the dataset. \cite{kane2022efficient} then approaches the VQA component of the dataset using basic combination strategies of the visual and textual domain features generated via modern vision and language architectures and training a multi-layer perceptron (MLP) on top of it. 

We use the data splits provided by \cite{kane2022efficient} for the VQA task, which consists of 4511 question-answering pairs across 1448 images. The training split consists of 3620 questions associated with 1158 images, and the testing split consists of 891 questions for 290 images. There are three types of questions: Yes/No, Counting and Condition Recognition. In this work, we use Yes/No, Image Condition Recognition and Road Condition Recognition questions to demonstrate the prowess of our methods.

\section{Results}
\label{app:permutations}
 
We report the results of our experiments on continual learning for VQA on the FloodNet datasets. We experiment on $3$ different episodic memory techniques for Experience Replay on all $6$ possible permutations of the task order. 
\begin{figure}
\begin{subfigure}{.5\textwidth}
  \centering
  \includegraphics[width=.8\linewidth]{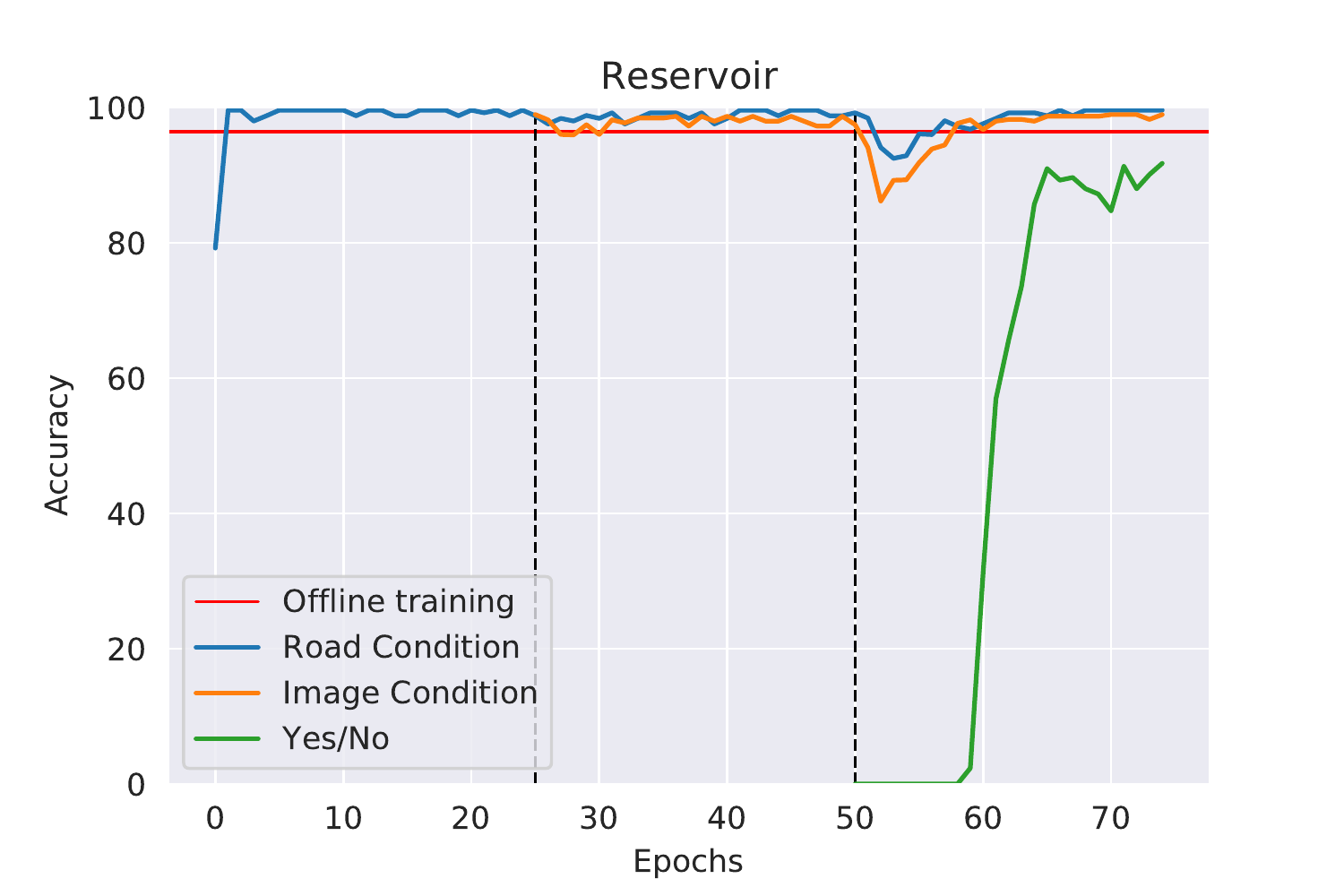}
\end{subfigure}%
\begin{subfigure}{.5\textwidth}
  \centering
  \includegraphics[width=.8\linewidth]{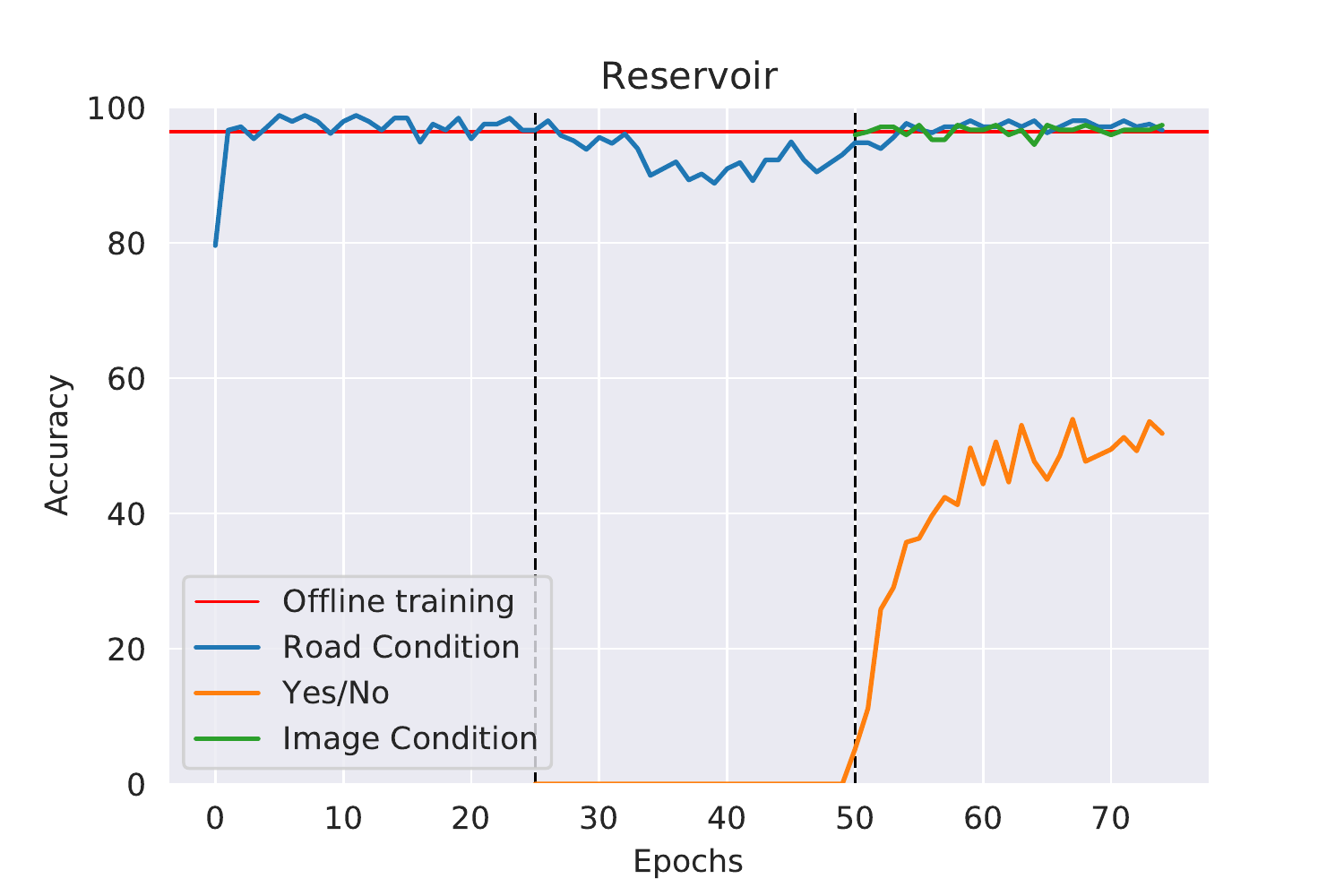}
\end{subfigure}
\\
\begin{subfigure}{.5\textwidth}
  \centering
  \includegraphics[width=.8\linewidth]{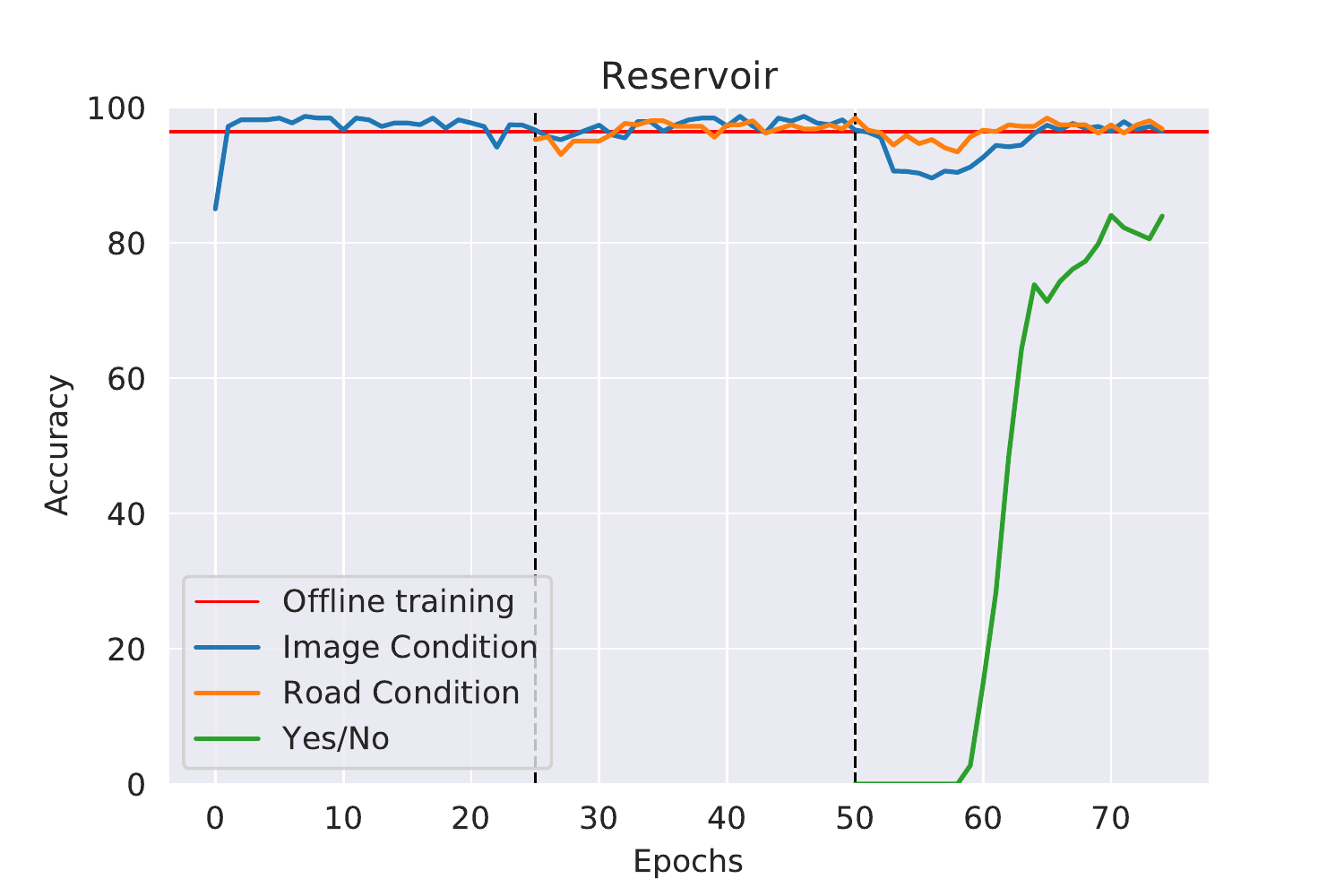}
\end{subfigure}%
\begin{subfigure}{.5\textwidth}
  \centering
  \includegraphics[width=.8\linewidth]{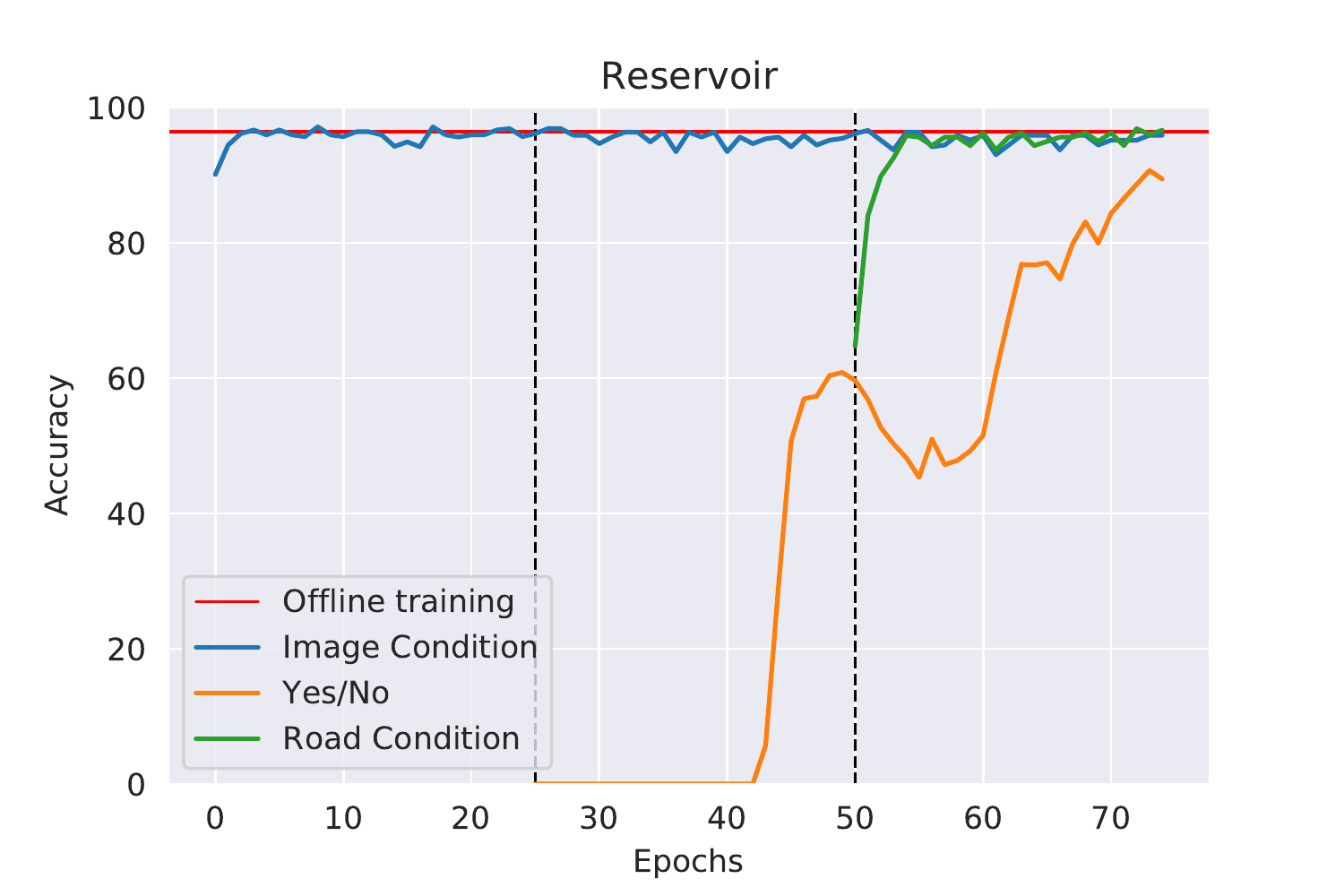}
\end{subfigure}%
\\
\begin{subfigure}{.5\textwidth}
  \centering
  \includegraphics[width=.8\linewidth]{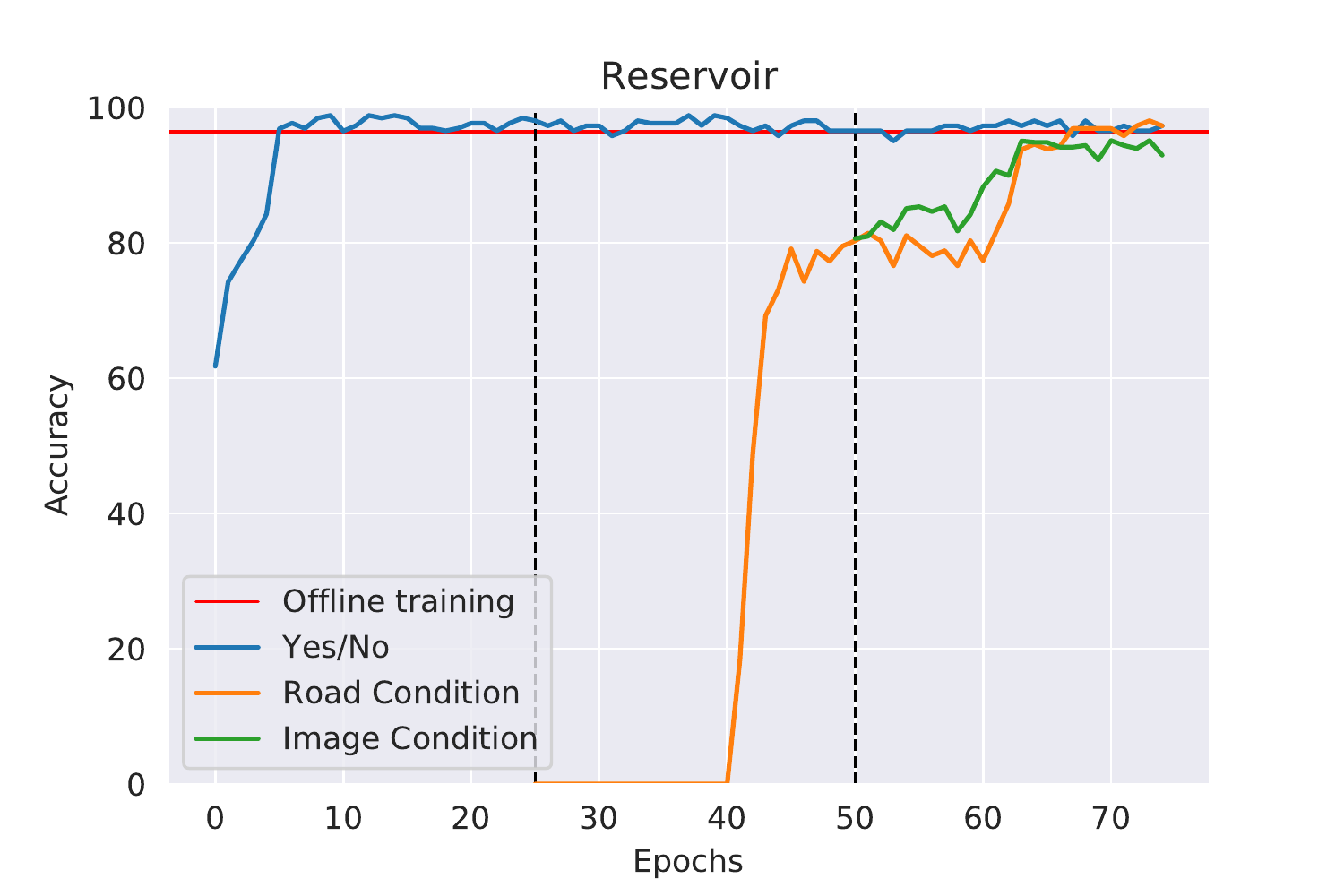}
\end{subfigure}%
\begin{subfigure}{.5\textwidth}
  \centering
  \includegraphics[width=.8\linewidth]{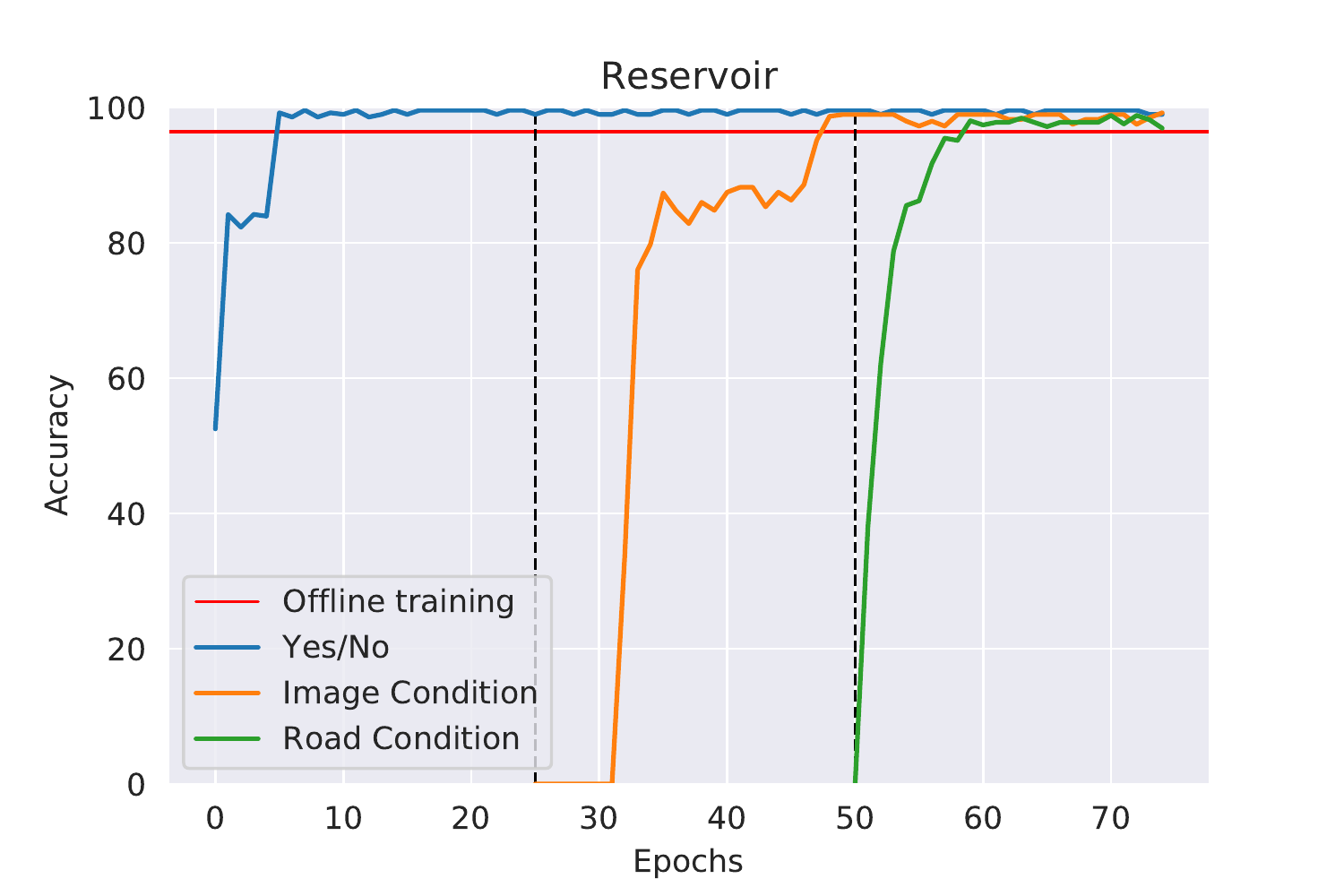}
\end{subfigure}%

\end{figure}

\begin{figure}
\begin{subfigure}{.5\textwidth}
  \centering
  \includegraphics[width=.8\linewidth]{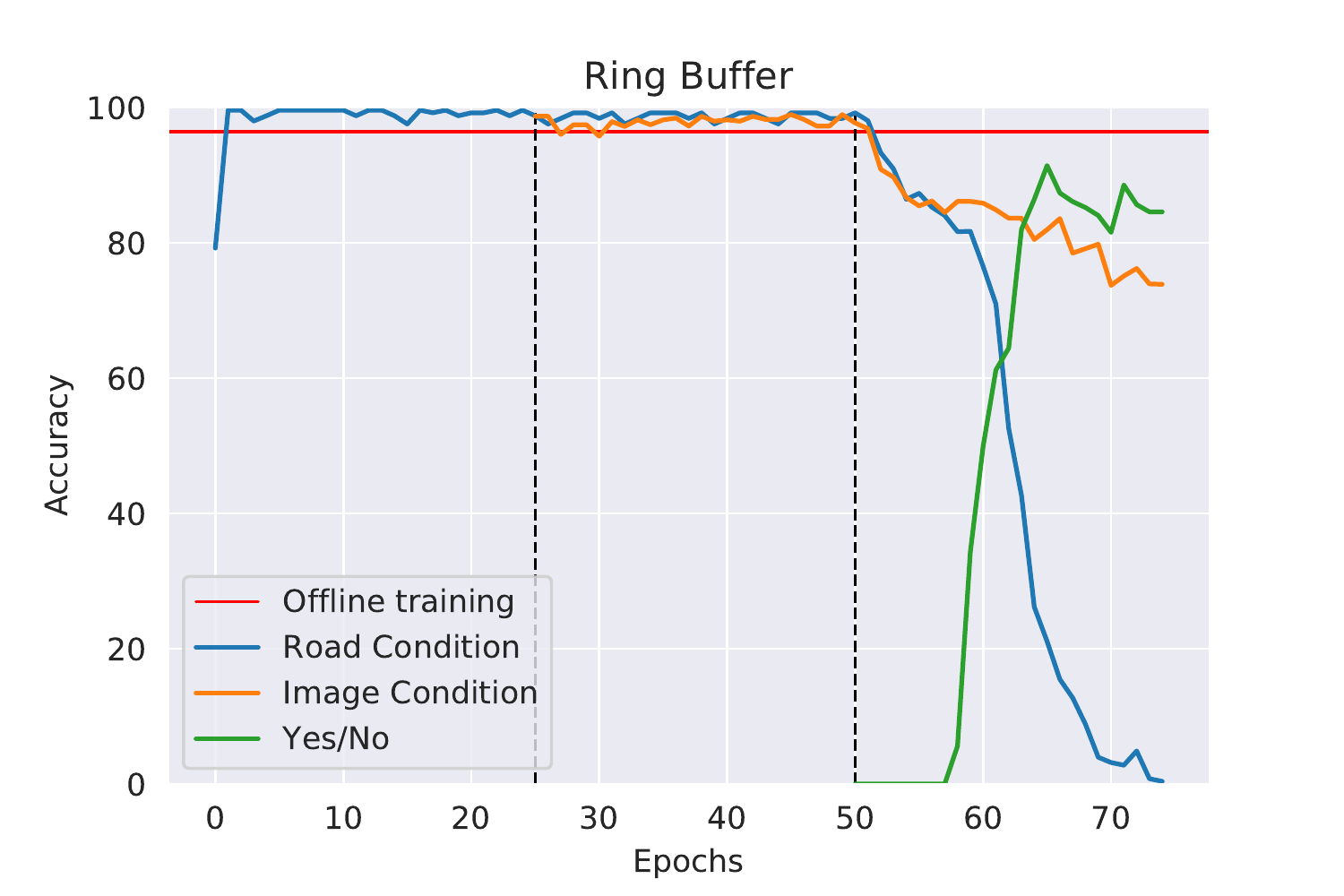}
\end{subfigure}%
\begin{subfigure}{.5\textwidth}
  \centering
  \includegraphics[width=.8\linewidth]{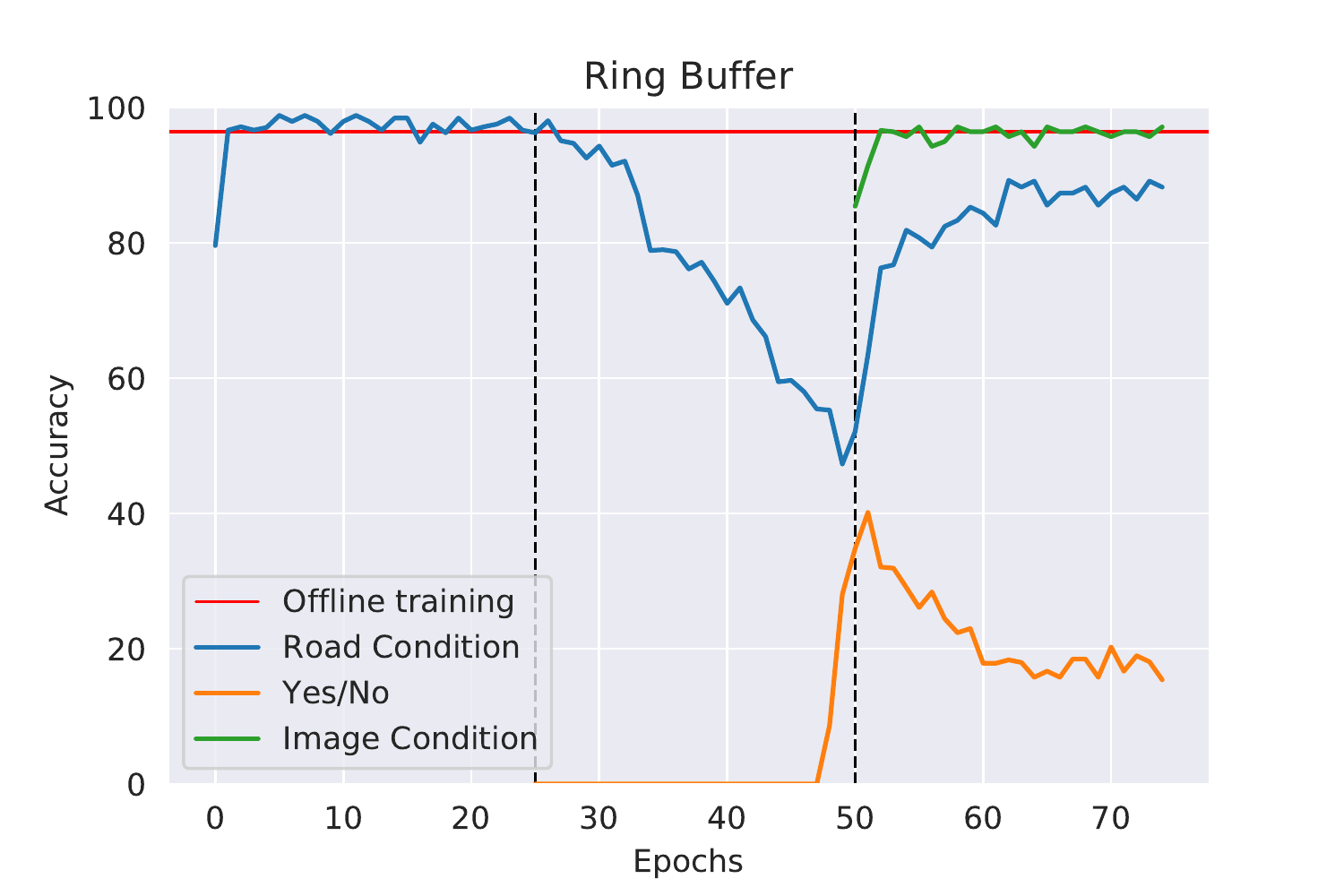}
\end{subfigure}
\\
\begin{subfigure}{.5\textwidth}
  \centering
  \includegraphics[width=.8\linewidth]{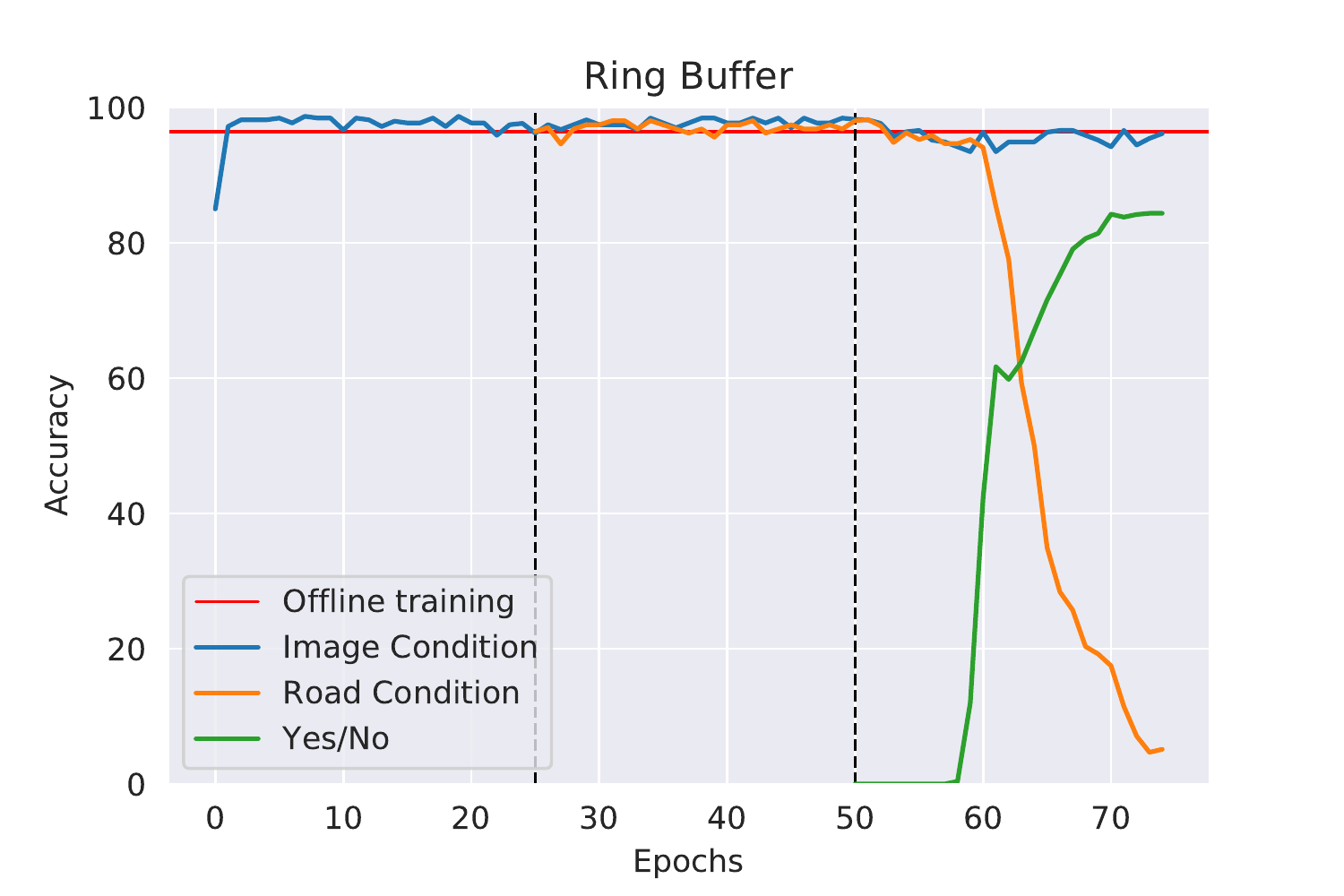}
\end{subfigure}%
\begin{subfigure}{.5\textwidth}
  \centering
  \includegraphics[width=.8\linewidth]{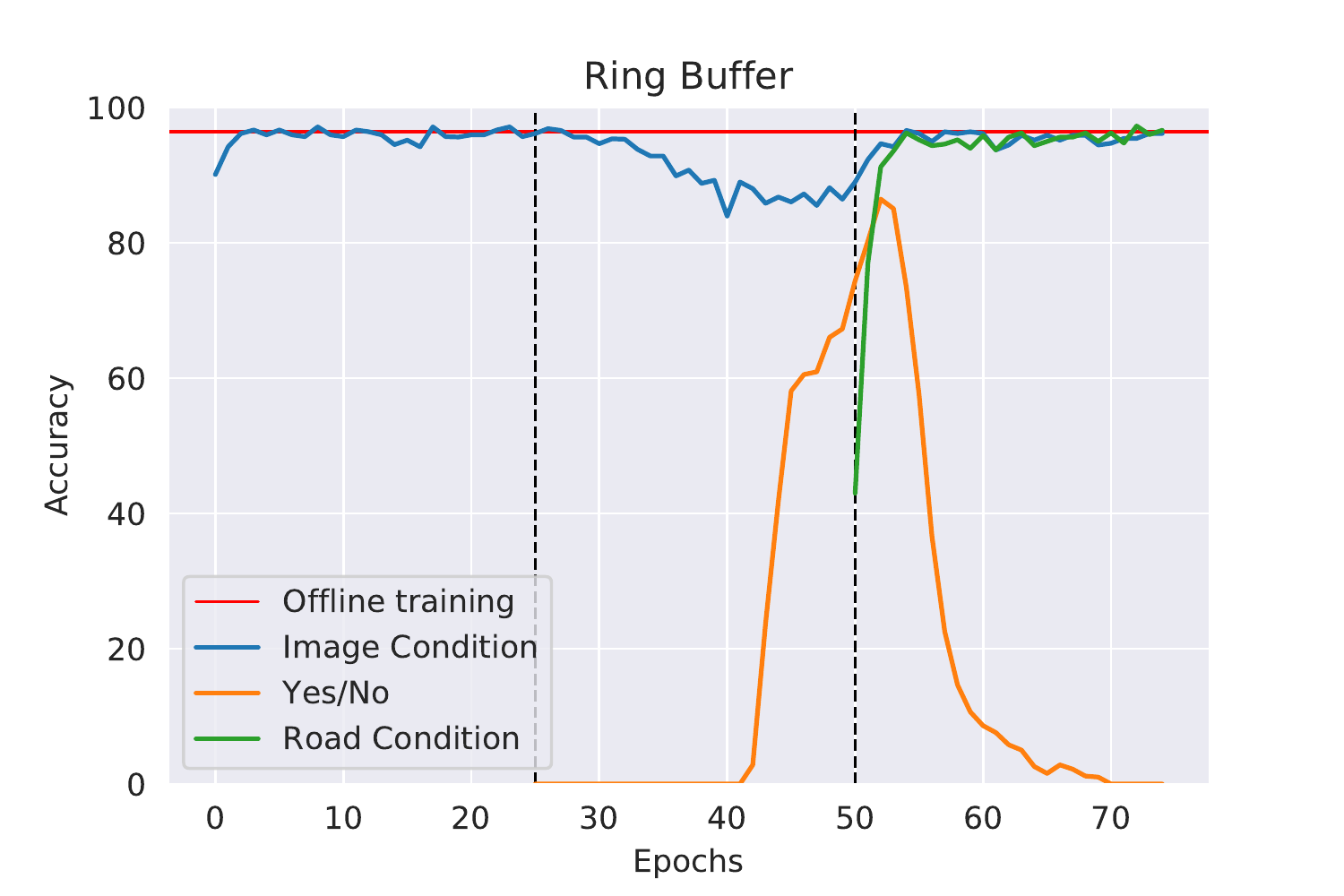}
\end{subfigure}%
\\
\begin{subfigure}{.5\textwidth}
  \centering
  \includegraphics[width=.8\linewidth]{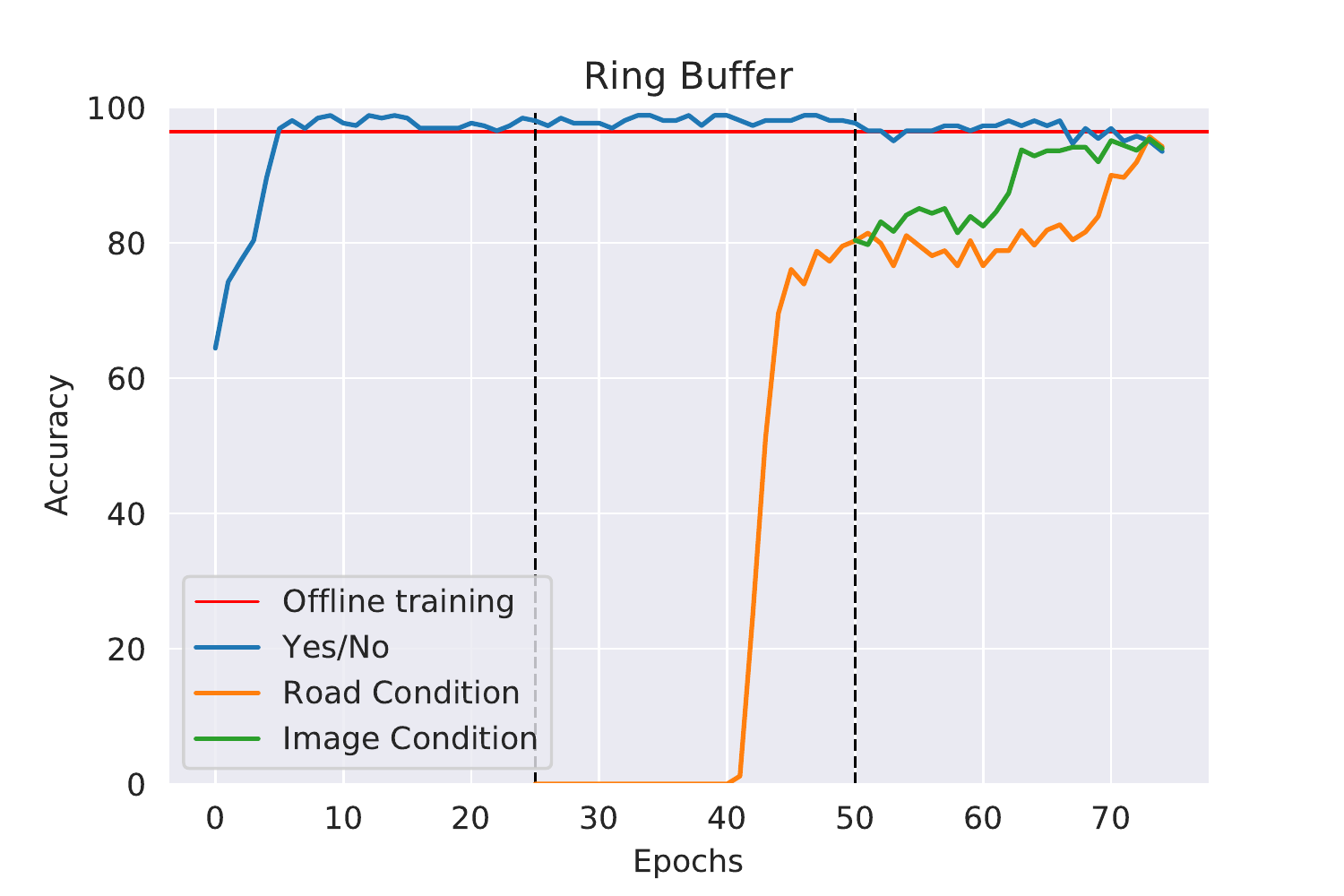}
\end{subfigure}%
\begin{subfigure}{.5\textwidth}
  \centering
  \includegraphics[width=.8\linewidth]{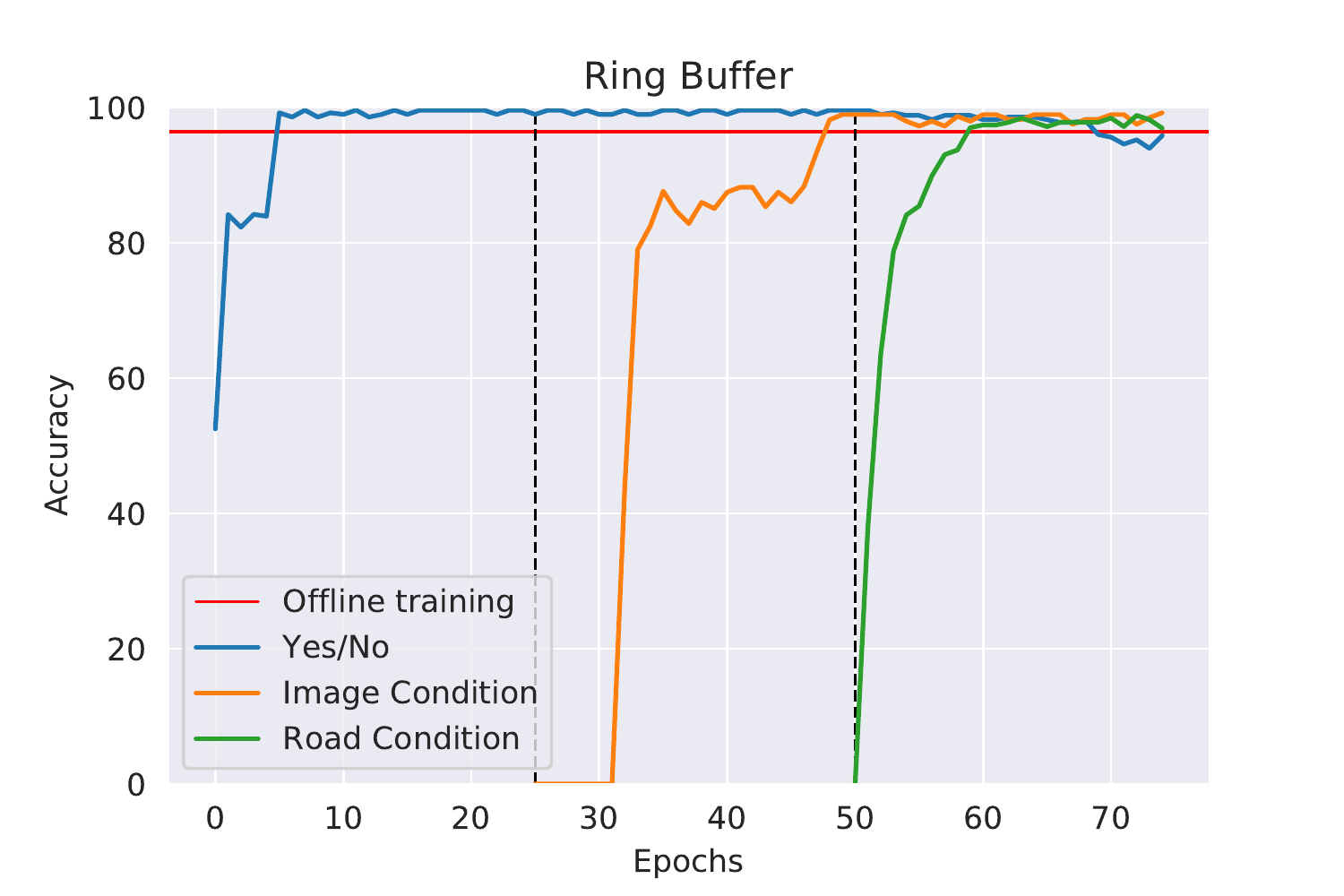}
\end{subfigure}%

\end{figure}

\begin{figure}
\begin{subfigure}{.5\textwidth}
  \centering
  \includegraphics[width=.8\linewidth]{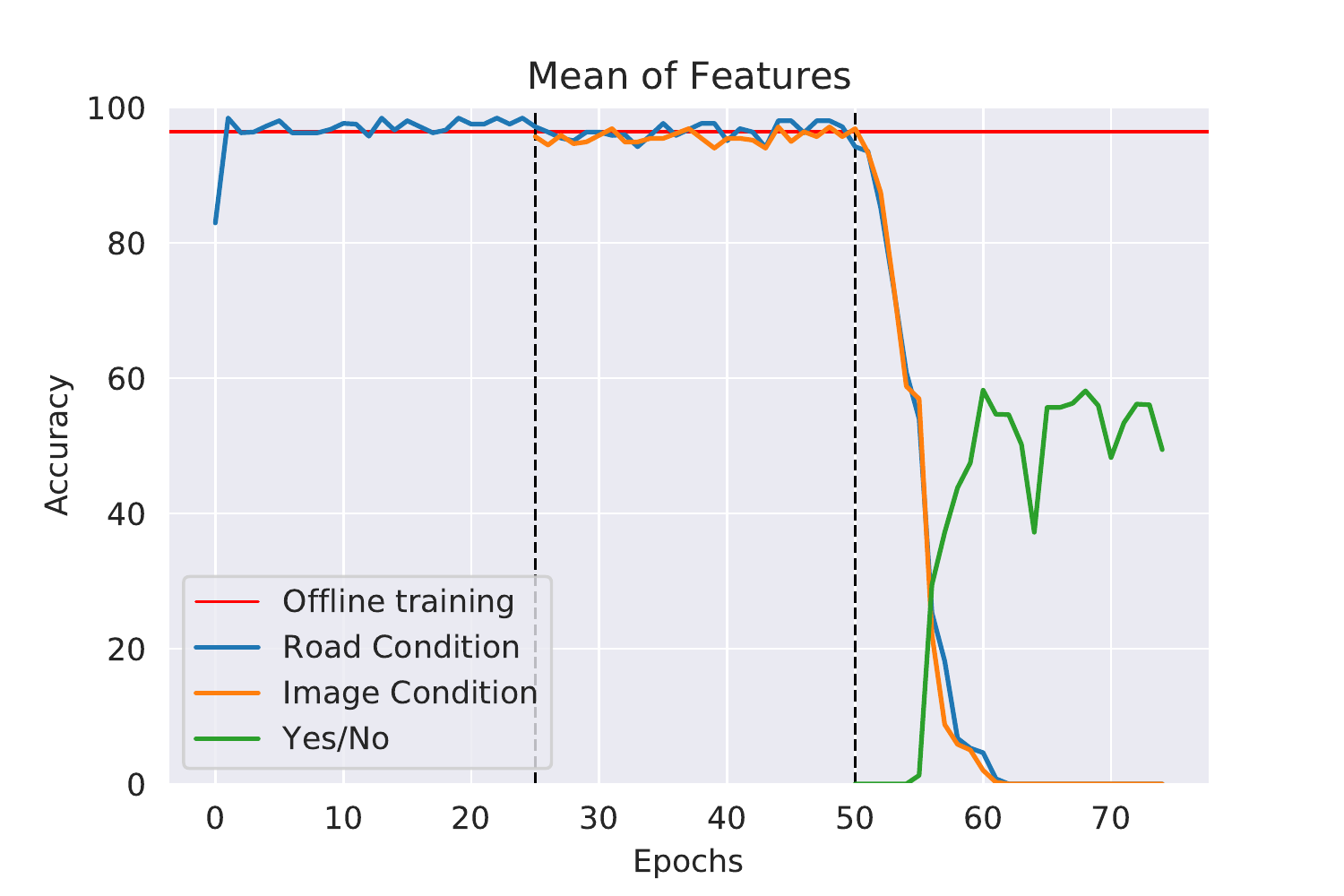}
\end{subfigure}%
\begin{subfigure}{.5\textwidth}
  \centering
  \includegraphics[width=.8\linewidth]{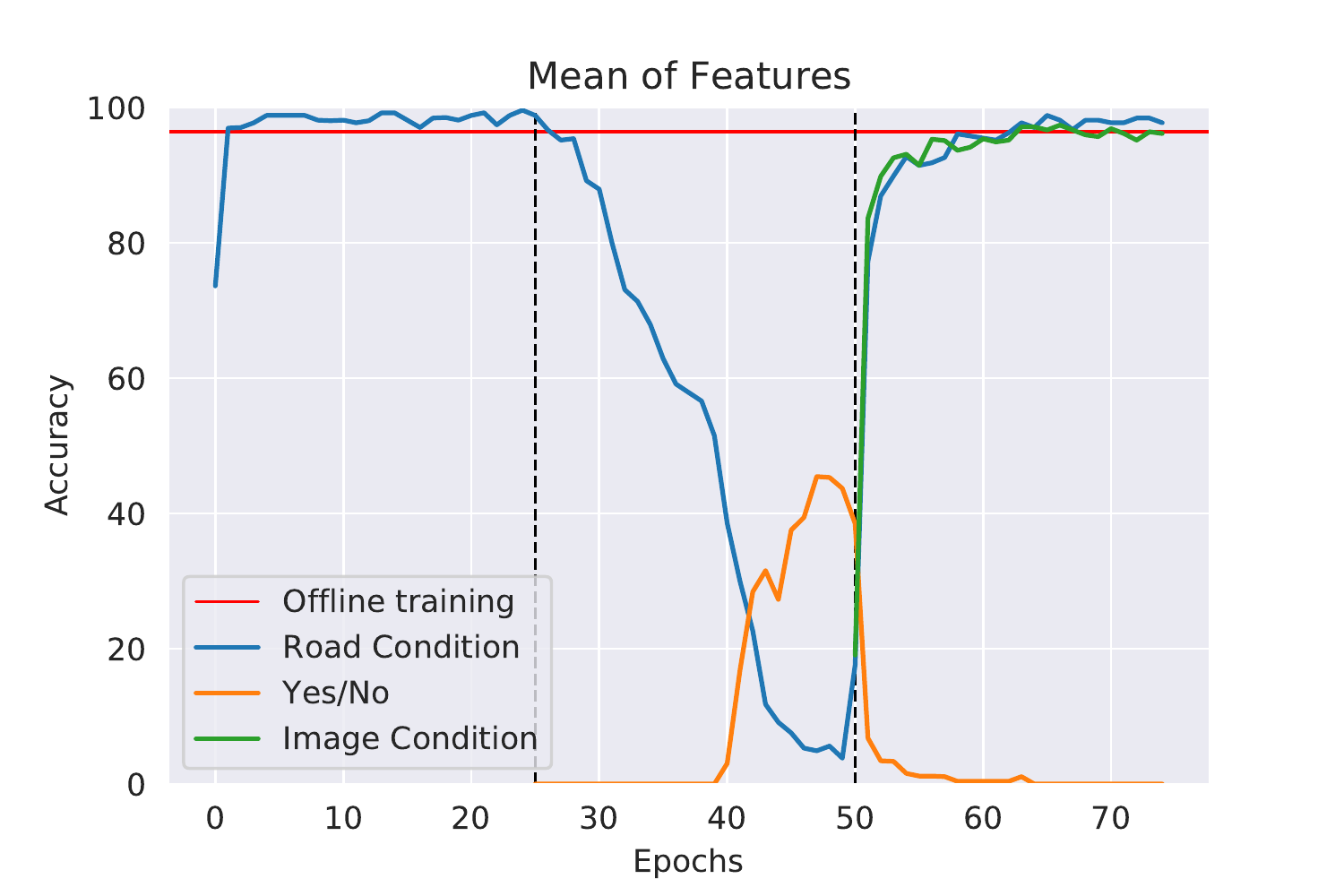}
\end{subfigure}
\\
\begin{subfigure}{.5\textwidth}
  \centering
  \includegraphics[width=.8\linewidth]{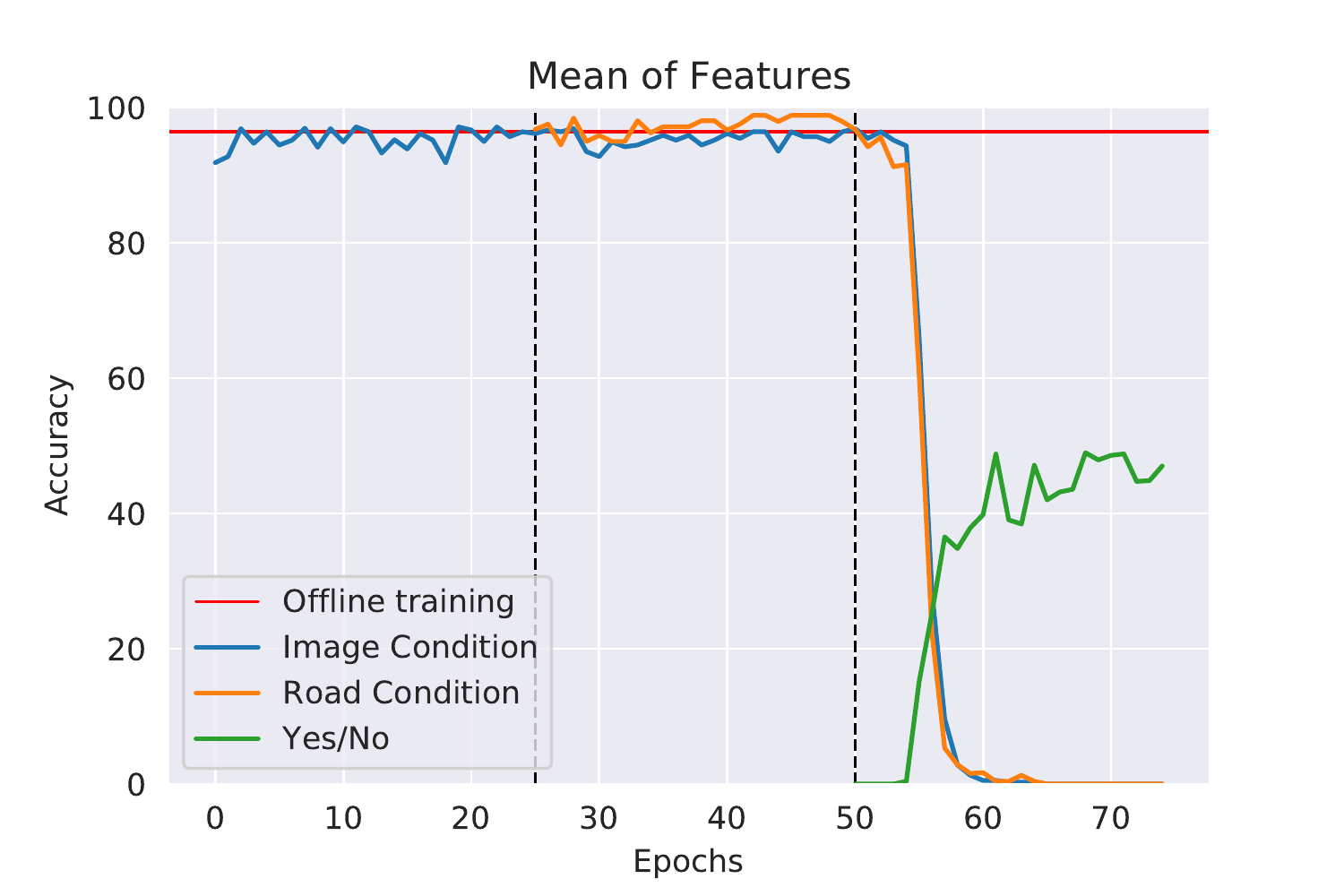}
\end{subfigure}%
\begin{subfigure}{.5\textwidth}
  \centering
  \includegraphics[width=.8\linewidth]{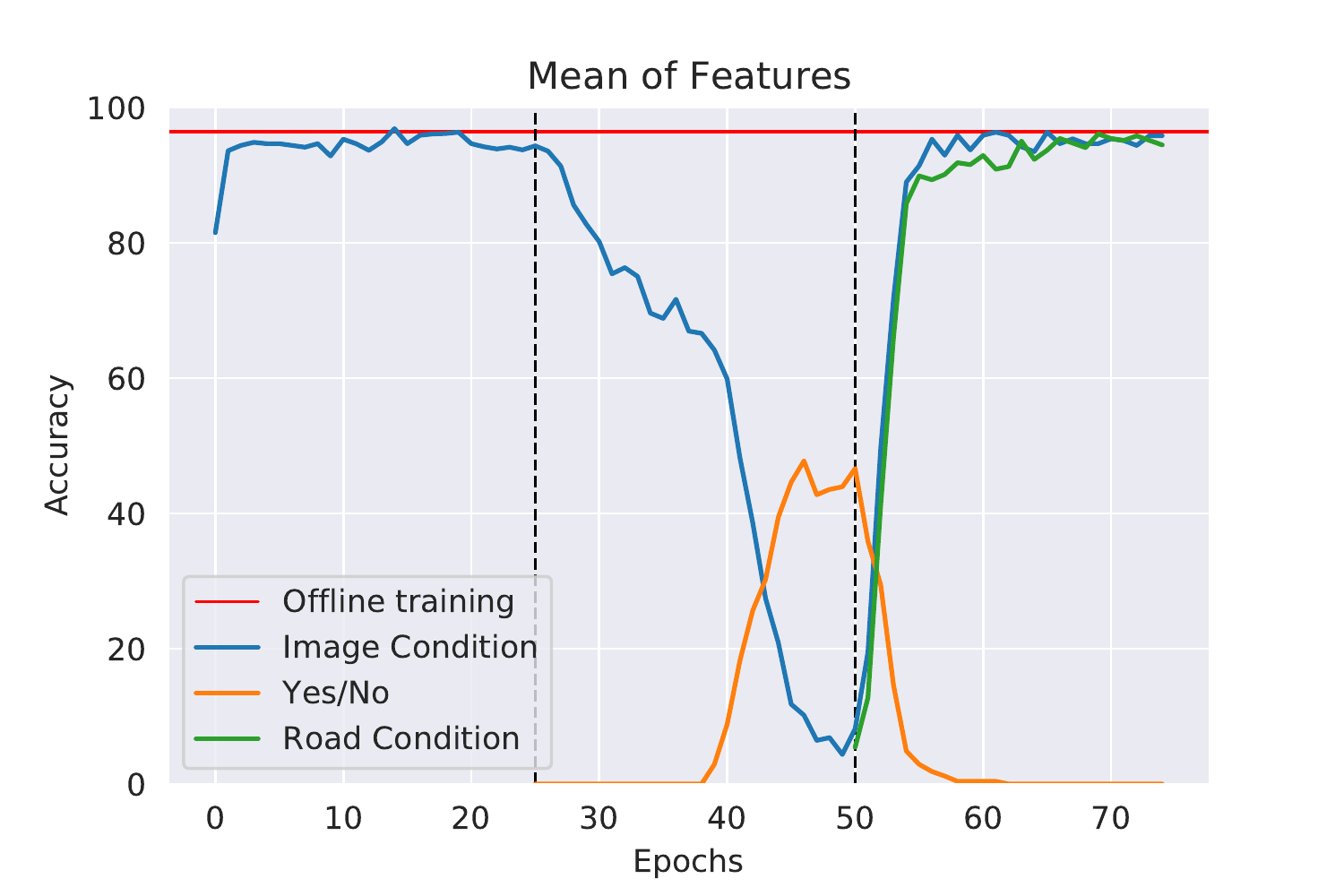}
\end{subfigure}%
\\
\begin{subfigure}{.5\textwidth}
  \centering
  \includegraphics[width=.8\linewidth]{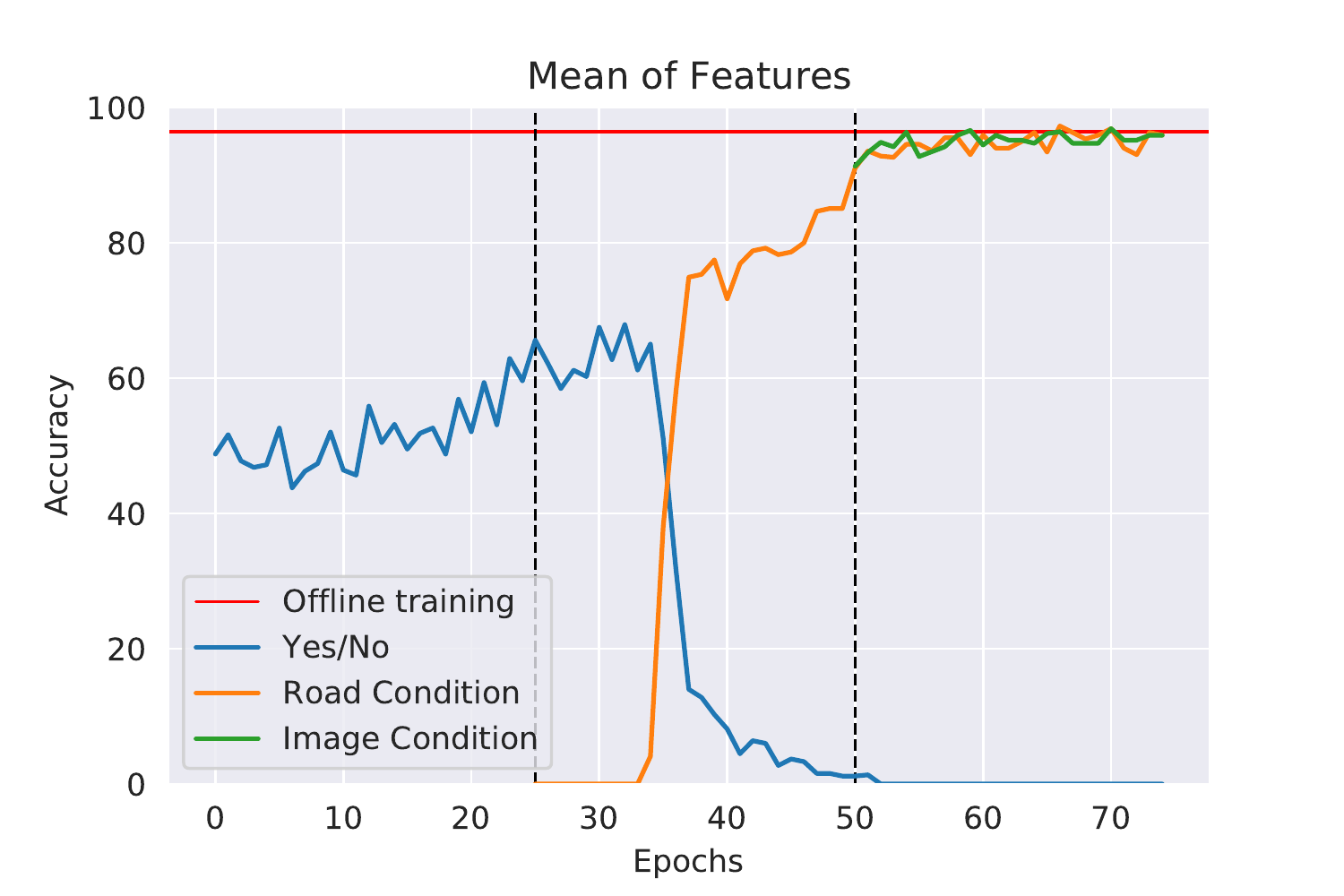}
\end{subfigure}%
\begin{subfigure}{.5\textwidth}
  \centering
  \includegraphics[width=.8\linewidth]{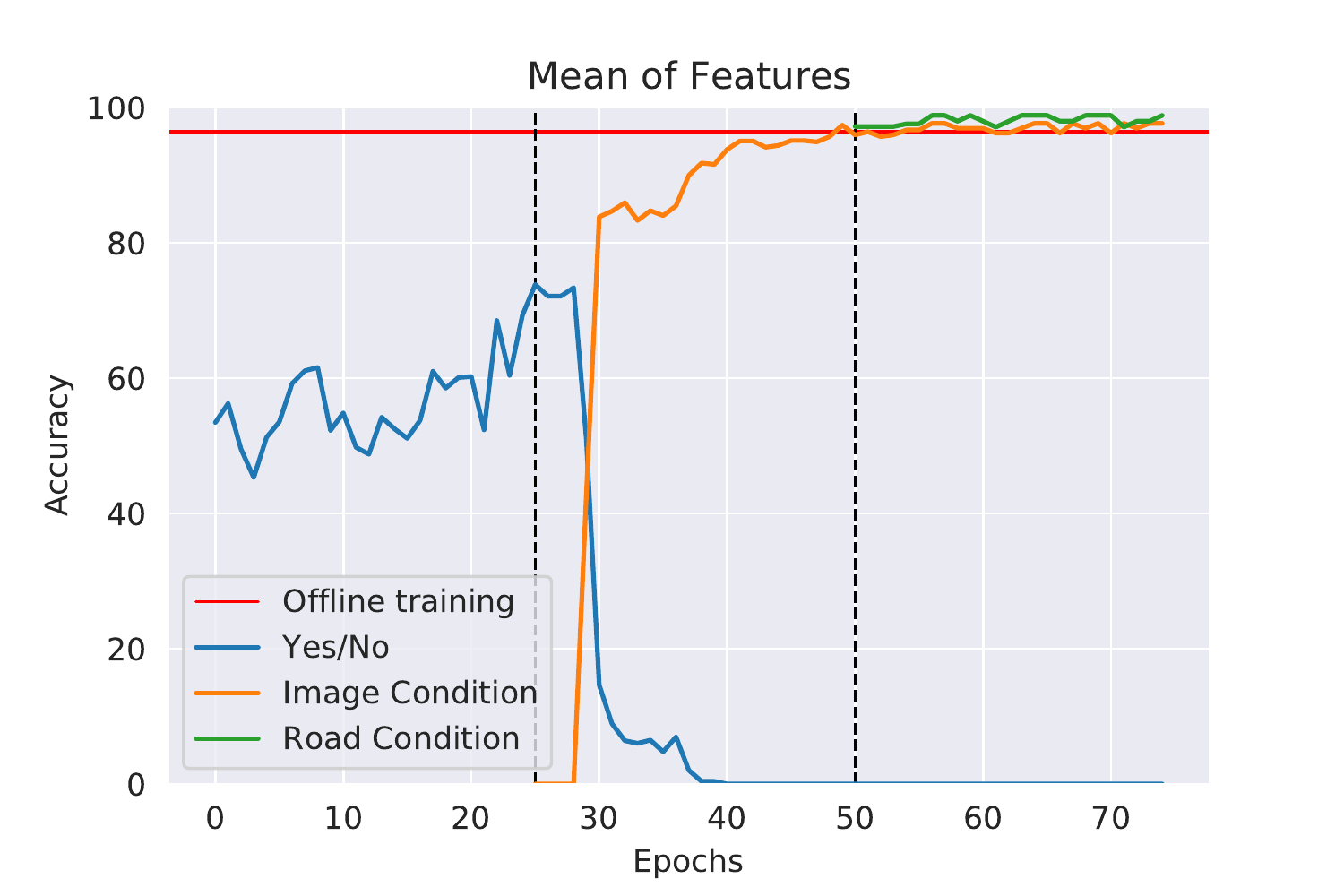}

\end{subfigure}%
  \caption{Our results on continual VQA setup using Experience Replay with three episodic memory techniques: Reservoir Sampling Update, Ring Buffer and Mean of Features and all possible task permutations. }
\end{figure}

\section{Implementation details }
\label{app:impl}
As mentioned earlier, we use experience replay for continual training. The basic method is as follows: we train the model on a task and very few samples from that task are stored in memory. In the later tasks, some samples are drawn from this memory and added to the training batch of the current task.

Our hyperparameters are as follows: we use a test split of $0.2$, batch size of $256$. We used LR of $1e-4$ for the first task and $5e-6$ for the others. We use a dropout rate of $0.2$. We use weight decay of $1e-5$ for first task and $2e-5$ for the later tasks. We use $25$ samples per class in our episodic memory. Finally, we train for $25$ epochs and our model has $1024$ hidden dimensions.



\end{document}